\documentclass[runningheads]{llncs}
\usepackage[T1]{fontenc}
\usepackage{framed,multirow}
\usepackage{cite}
\usepackage{amsmath}
\usepackage{amssymb}
\usepackage{graphicx,verbatim}
\usepackage{utfsym}
\usepackage[misc]{ifsym}

\begin{document}
\title{Information Bottleneck-based Causal Attention for Multi-label Medical Image Recognition}
%

\author{Xiaoxiao Cui\inst{1}\and 
Yiran Li\inst{2} \and 
Kai He\inst{2} \and
Shanzhi Jiang\inst{2} \and 
Mengli Xue\inst{2} \and 
Wentao Li\inst{1} \and
Junhong Leng\inst{4} \and
Zhi Liu$^{(\textrm{\Letter})}$\inst{2} \and 
Lizhen Cui$^{(\textrm{\Letter})}$\inst{1} \and 
Shuo Li\inst{5} } 
\authorrunning{X. Cui et al.}
%
\institute{Joint SDU-NTU Centre for Artificial Intelligence Research (C-FAIR), Shandong University, Jinan, Shandong 250101, China \\ \and
School of Information Science and Engineering, Shandong University, Qingdao, Shandong 266237, China \\
\email{\{clz, liuzhi\}@sdu.edu.cn} \and
Department of ultrasound, Jinan Maternity and Child Care Hospital Affiliated to Shandong First Medical University, Jinan, Shandong 250012, China \and
Case Western Reserve University, Cleveland, OH 44106, USA \\
}
\maketitle              

\begin{abstract}
Multi-label classification (MLC) of medical images aims to identify multiple diseases and holds significant clinical potential. A critical step is to learn class-specific features for accurate diagnosis and improved interpretability effectively. However, current works focus primarily on causal attention to learn class-specific features, yet they struggle to interpret the true cause due to the inadvertent attention to class-irrelevant features. To address this challenge, we propose a new structural causal model (SCM) that treats class-specific attention as a mixture of causal, spurious, and noisy factors, and a novel Information Bottleneck-based Causal Attention (IBCA) that is capable of learning the discriminative class-specific attention for MLC of medical images. Specifically, we propose learning Gaussian mixture multi-label spatial attention to filter out class-irrelevant information and capture each class-specific attention pattern. Then a contrastive enhancement-based causal intervention is proposed to gradually mitigate the spurious attention and reduce noise information by aligning multi-head attention with the Gaussian mixture multi-label spatial. Quantitative and ablation results on Endo and MuReD show that IBCA outperforms all methods. Compared to the second-best results for each metric, IBCA achieves improvements of 6.35\% in CR, 7.72\% in OR, and 5.02\% in mAP for MuReD, 1.47\% in CR, and 1.65\% in CF1, and 1.42\% in mAP for Endo.
\end{abstract}
\section{Introduction}
\begin{figure}[h]
\includegraphics[width=\textwidth, trim=0 0 0 0, clip]{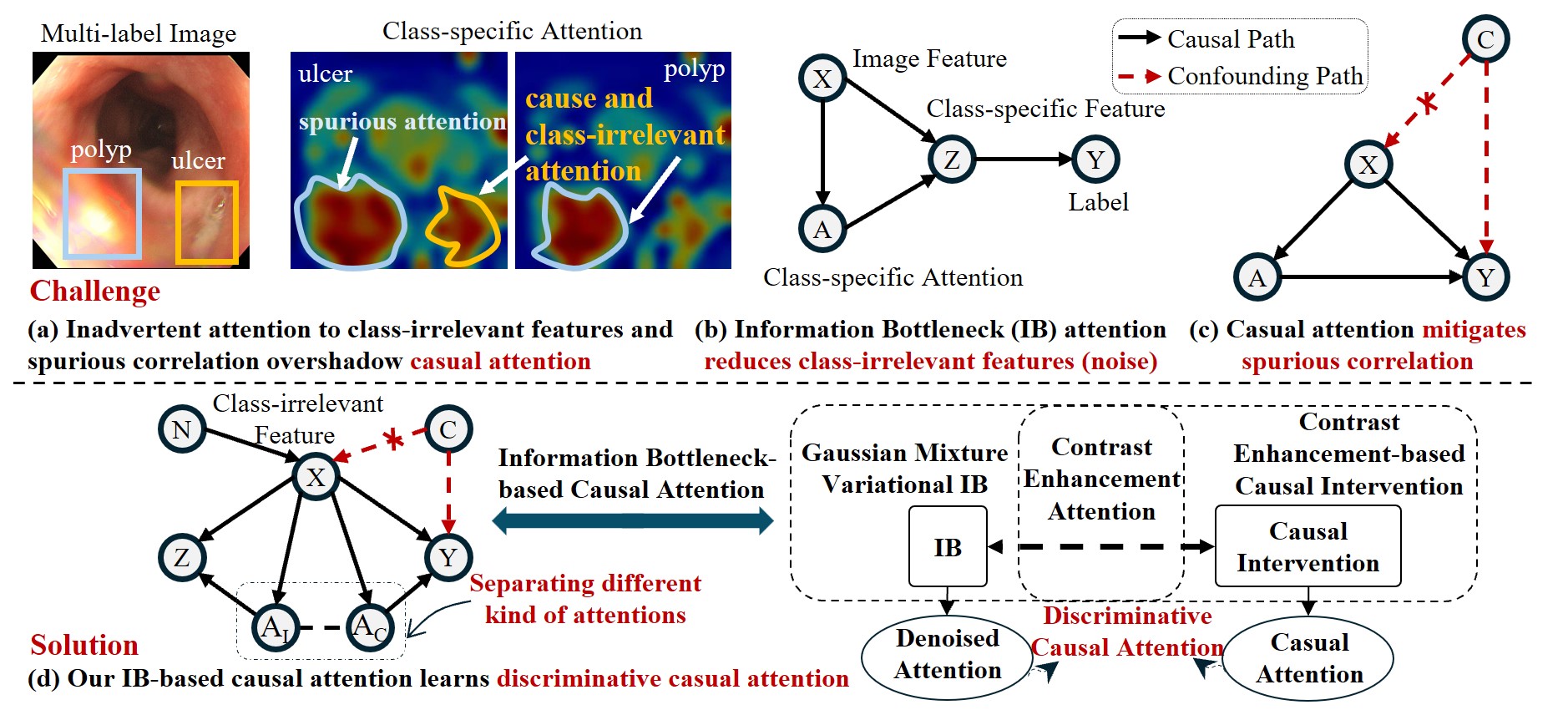}
\caption{Challenge and our solution. (a) The task-irrelevant features and spurious attention in spatial attention hinder causal class-specific patterns for MLC of medical images. While Information Bottleneck (IB) attention cannot mitigate the confounder factor (b) and causal attention cannot reduce noise information (c), we propose IB-based causal attention to address this challenge (d).}
\label{fig1}
\end{figure}
Multi-label classification (MLC) of medical images is crucial for the clinical diagnosis of diseases, as multiple diseases or conditions can co-occur within a single image~\cite{wang2023real,rodriguez2022multi,li2021benchmark,scholkopf2021toward}. Instead of training separate classifiers for each label, MLC seeks to learn a single model that predicts multiple labels simultaneously for a given input, thus promoting more efficient training. Therefore, it is crucial to learn comprehensive representations and effectively separate them into class-specific and class-irrelevant within each instance for accurate predictions.

Attention has demonstrated its effectiveness in learning class-specific representations to improve classification accuracy in MLC tasks. By capturing different regions~\cite{zhu2021residual,zhou2023aligning} or semantics~\cite{zhao2021transformer,zhu2024semantic} occupied by objects from various categories, attention facilitates the learning of robust class-specific features for MLC. However, attention cannot always access the causal factors and may incorrectly focus on the wrong context regions when training samples are insufficient~\cite{liu2022causality}. This contextual bias is considered confounder, leading to spurious correlations. This can be mitigated by causality learning~\cite{liu2022causality,cui2024multilevel} by highlighting the underlying causal regions in an image for more robust class-specific features.

Despite promising results of causality-based attention for class-specific feature learning, they still struggle to minimize redundancy information to ensure the most discriminative causal attention for each class. Redundancy is introduced when the attention mechanism captures class-irrelevant features, such as background and inherent anatomical structures in medical images, reducing its ability to interpret the cause. For example, polyps and ulcers exhibit similar textures under varying lighting conditions (Fig. \ref{fig1} (a)). This results in both causal factors and class-irrelevant features being incorporated into the attention, which inevitably introduces noise. While the Information Bottleneck (IB) principle~\cite{saxe2018information} can be applied to restrict information bypassed by attention~\cite{lai2021information}, it overly relies on label-related attention and ignores its causality. This limitation motivates the incorporation of IB into causal attention mechanisms. However, applying a spherical Gaussian prior to the latent class-specific features reduces the model’s ability to differentiate between multi-label class-specific attentions in MLC tasks.

To address this challenge, a new structural causal model (SCM) is constructed to regard class-specific attention as a mixture of causal, spurious, and noisy factors, and Information Bottleneck-based Causal Attention (IBCA) is proposed to learn discriminative causal class-specific attention for MLC of medical images (Fig. \ref{fig1} (d)). Specifically, class-irrelevant (noise) information is eliminated by Gaussian mixture variational IB to learn class-specific attention, which is incorporated into contrastive enhancement-based causal intervention to mitigate spurious and noise attention for discriminative causal class-specific features. Building upon this SCM, our \textbf{contributions} are: (1) For the first time, IBCA incorporates IB into causality learning for MLC of medical images, effectively separating spurious and noisy information from the causal class-specific attention. (2) IBCA advances causality learning by gradually incorporating Gaussian mixture multi-label spatial attention into causal intervention through contrastive learning. This enables the learning of discriminative causal attention with enhanced interpretability.
\section{Information Bottleneck-based Causal Attention}
\subsection{Causal Inference for MLC of Medical Images}
\textbf{Attention Information Bottleneck.} For MLC tasks, the image feature $X$, class-specific attention $A$, latent class-specific feature $Z$, and prediction $Y$ follow the joint conditional distribution of the model shown in Fig. \ref{fig1} (b). The mutual information between $Z$ and $Y$ is $I(Z; Y) = \int_{z \in Z} \int_{y \in Y} p(z, y) \log \frac{p(z, y)}{p(z)p(y)} \, dz \, dy$. To remove label-irrelevant information in $A$, IB is applied to optimize the objective: $\min_{\substack{A}} -I(Z; Y) + \beta I(Z; A, X)$, where $\beta >0 $ controls the trade-off between two MI terms, encouraging $A$ to learn minimal redundancy attention that is most predictive of $Z$. Based on variational approximation~\cite{alemi2016deep}, the lower bound of the attentive variational information bottleneck can be obtained by:
\begin{equation}
    \mathcal{L}_{VIB} = I(Z; Y) - \beta I(Z; A, X).
    \label{eq0}
\end{equation}
The first term is the negative loglikelihood of the prediction, aiming to learn $Z$ that can maximally obtain the correct prediction. The second term aims to minimize the MI between $X$ and $Z$ with given $A$.
\\
\textbf{Causal Attention.} An SCM is built as a causal view to describe the causality among four variables in MLC of medical images: $X$, $A$, $Y$, and confounder $C$ in Fig. \ref{fig1} (c). $C$ influences $X$, which in turn affects $A$, introducing a spurious correlation into the desired causal effect: $X \rightarrow A \rightarrow Y$, resulting in spurious correlations: $C \rightarrow X \rightarrow A \rightarrow Y$. To eliminate the confounding effect caused by $C$, causal intervention cuts off the links from $C$ to $X$ and $A$ to identify the causal attention. This is implemented by a backdoor adjustment with do-calculus~\cite{pearl2016causal}: $\mathrm{P}(Y \mid \text{do}(X))$. To approximate the confounding effects, sampling all possible confounding impacts can be approximated by sampling $N$ times on the observed data $(x,y)$~\cite{pearl2009causal}. In this way, different class-specific attention $a_k^n$ for class $k$ of sample $n$ is obtained, and the causal intervention can be modeled as the sigmoid activated classification probability of the class-specific features:
\begin{equation}
\begin{split}
P(Y = k | do(X = x)) &= \sum_{n=1}^{N} \frac{P(Y = k | X = x^n) P(X = x^n)}{P(X = x^n | C = c)} \\
&= \sum_{n=1}^{N} \frac{\text{Sigmoid}(\text{Clf}(a_k^nx^n)) P(a_k^n)}{P(a_k^n| c)},
\end{split}
\label{eq1}
\end{equation}
where $\text{Clf}(\cdot)$ denotes the classifier corresponding to each class $k$, and $a_k^n x^n$ represents the class-specific features. The term $\frac{P(a_k^n)}{P(a_k^n \mid c)}$ indicates the weight of each spatial attention sample, which is set to $1/N$ due to uniform sampling in the multi-head attention mechanism. Further implementation details are provided in Section 2.3.
\\
\textbf{Information Bottleneck-based Causal Attention.} We construct an SCM of IB-based causal attention by incorporating noise $\mathbf{N}$ into the causal attention mechanism (Fig. \ref{fig1} (d)). Although $\mathbf{N}$ introduces no spurious correlation with $Y$, it mitigates the discriminative ability of causal attention $A_c$. We propose explicitly separating attention into noise, causal, and spurious components, and design a simple IBCA approach that introduces contrastive enhancement attention with IB and causal intervention. Details will be provided in the following two sections.
\begin{figure}[t]
\includegraphics[width=\textwidth, trim=0 0 0 0, clip]{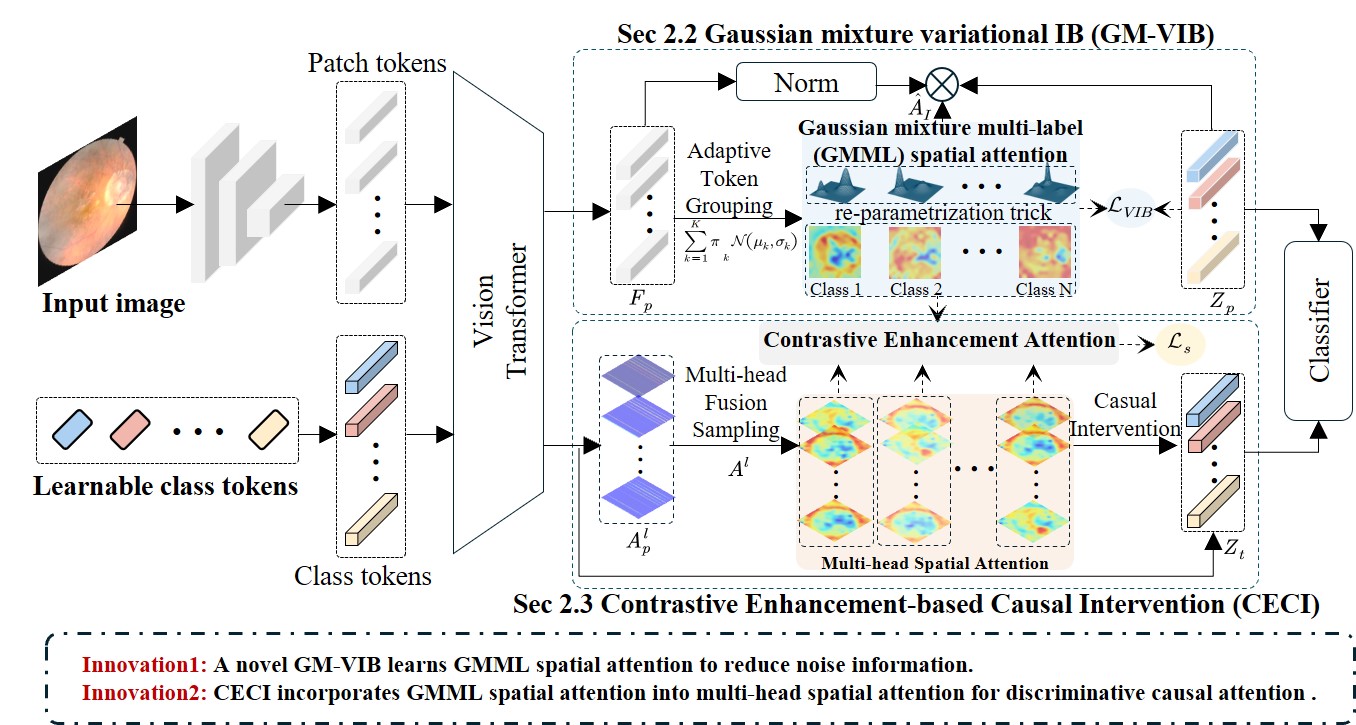}
\caption{Overview of our IBCA model, which is built on a transformer to learn Gaussian mixture multi-label spatial attention for noise reduction and is further integrated with contrastive enhancement-based causal intervention to achieve discriminative causal attention learning.}
\label{fig2}
\end{figure}
\subsection{Gaussian Mixture Variational Information Bottleneck}
Our Gaussian Mixture Variational Information Bottleneck (GM-VIB) learns Gaussian mixture multi-label (GMML) spatial attention to effectively filter out class-irrelevant information. The distribution parameters of each Gaussian component are learned by adaptive token grouping and the VIB constraint.
\\
\textbf{Adaptive Token Grouping.} Given an input image $x$, its multi-label spatial attention maps $A_I$ are derived from the patch tokens $F_p$ learned by a multi-class token transformer~\cite{xu2024mctformer+}. Each spatial attention map follows a Gaussian mixture distribution, with the number of Gaussian components equal to the number of multi-label classes $N_c$. For each Gaussian component, the reparameterization trick~\cite{alemi2016deep} is used to calculate its mean vector ${\mu}_k$, covariance matrix ${\sigma}_k$, and coefficient ${\pi}_k$. This is implemented by applying three different multiple liner projection layers to $F_p$. In this way, the Gaussian mixture representation is learned as $ \sum_{k=1}^{K} {\pi}_k \mathcal{N}(\mu_k, {\sigma}_k),k \in \{1,2,\cdots, N_c\}$. ${\pi}_k$ satisfies $0 \leq {\pi}_k \leq 1, \sum_{k=1}^{N_c} {\pi}_k = 1$. 
\\
\textbf{Variational Information Bottleneck.} To ensure that the learned spatial attention is class-specific, different Gaussian mixture multi-label spatial attention samples are obtained by resampling from the Gaussian mixed distribution to derive $Z$ in Eq. \ref{eq0}. Specifically, the attention samples $\hat{A_I}$ are obtained by sampling a new set of $\Pi=\{\pi_1,\pi_2, \cdots, \pi_{K}\}$ from a Gamma distribution for each class~\cite{henderson2023vae}, and sampling each vector independently from a component of the prior Gaussian distribution ${G}^p$. The class-aware features $Z_p$ are then learned by multiplying $F_p$ with $\hat{A_I}$ as: $Z_p=Norm(F_p) \times \hat{A_I}, Z_p \in \mathbb{R}^{C \times D}$, where $Norm(\cdot)$ denotes normalization operation. To produce class scores, $Z_p$ is passed through a Global Average Pooling (GAP) layer, which is supervised by a Multi-Label Soft Margin (MLSA) loss. Thus, the first term in Eq. \ref{eq0} can be rewritten as ${L}_{mlsm}\{GAP(Z_p),y\}$. We set ${G}^p$ as a fixed spherical Gaussian: ${G}^p = \mathcal{N}(0, 1)$, and the second term in Eq. \ref{eq0} can be substituted by calculating the Kullback-Leibler (KL) divergence between the approximate posterior ${G}_p$ and the true posterior $\mathcal{N}(\mu_k, {\sigma}_k)$ by:
\begin{equation}
{KL}({G}_k^q\parallel {G}^p)=\frac{1}{2}\sum_{k=1}^{N}(\mu_{k}^2+\sigma_{k}^2-\text{log}(\sigma_{k})-n).
\label{eq2}
\end{equation}
We assume $n = 1$ to simplify the derivation process. Eq. (\ref{eq0}) is rewritten as: $\mathcal{L}_{VIB} = {L}_{mlsm}\{GAP(Z_p),y\} + \beta {KL}({G}_k^q\parallel {G}^p)$. This constrains spatial attention by reducing noise to highlight the most contributing regions for each class.
\subsection{Contrastive Enhancement-based Causal Intervention}
Our Contrastive Enhancement-based Causal Intervention (CECI) gradually incorporates GMML spatial attention into Multi-Head Fusion Sampling (MHFS)-based causal intervention, constraining noise information to enable discriminative causal attention by a Contrastive Enhancement Attention (CAE) loss.
\\
\textbf{MHFS-based Causal Intervention.} Multi-head self-attention is incorporated into the causal intervention due to the embedded class-specific information in it for a multi-class token transformer. To generate class-specific attentions of MHFS, for each head self-attention map $A^{l}_p \in \mathbb{R}^{(N_p^2+N_c) \times (N_p^2+N_c)}$ from the last transformer block, the class-to-patch attention $A^{l}_{c2p} = A^{l}_p[1:N_c, N_c +1: N_c + N^2]$ and patch-to-patch affinity $A^{l}_{p2p} = A^{l}_p[N_c +1:N_c+ N_p^2, N_c +1: N_c + N_p^2]$ are multiplied to generate a class-specific attention sample $A^{l} \in \mathbb{R}^{N_c \times N_p^2}$, where $N_p$ is the patch size. Instead of directly applying each class-specific attention sample to the classification to ensure causality, all token embeddings are incorporated into the classification~\cite{xu2024mctformer+}. Since patch-tokens are incorporated in the classification in the VIB loss, a GAP is applied to multiple class-tokens $F_t$ to learn a class-specific feature $Z_t$, with MLSA loss $L_{t}$ to constrain the training.
\\
\textbf{CAE} aligns each $A^l$ with Gaussian mixture spatial attention to reduce noise information in the causal intervention. Specifically, each attention is activated by a sigmoid function, and the pairwise cosine similarity between $A^{l}$ and $\hat{A_I}$ is calculated and averaged across all heads to formulate the CAE loss $\mathcal{L}_{s}$ by:
\begin{equation}
\mathcal{L}_{s} = \frac{1}{H} \sum_{l=1}^{H} \mathcal{L}_{s}^{l},
\quad \mathcal{L}_{s}^{l} =1- \frac{\hat{A}_I A^{l}}{\|\hat{A}_I\| \|A^{l}\|}.
\label{eq3}
\end{equation}
$H$ is the number of attention heads. This forces each $A^l$ to receive stronger and more frequent guidance to learn discriminative causal class-specific information.

The final objective to minimize is simply the summation of different losses: $\mathcal{L}=\mathcal{L}_{VIB}+\mathcal{L}_{t}+\lambda_{s}\mathcal{L}_{s}$, where $\lambda_{s}$ is the trade-off weights and we set $0.01$ here.
\section{Experiments and Results}
\subsection{Dataset and Implementation Details}
\textbf{Dataset and Evaluation.} We conduct experiments on two publicly available datasets: Endo~\cite{wang2023real} and MuReD~\cite{li2021benchmark}. Endo consists of 3865 colonoscopy images, annotated with 4 types of lesions. MuReD includes 2,208 samples collected from three different sources (ARIA~\cite{farnell2008enhancement}, STARE~\cite{hoover2000locating}, and RFMiD~\cite{pachade2021retinal}), covering a wider range of diseases with 20 label types. Both datasets are randomly split into training, validation, and test sets in an 8:1:1 ratio.

In addition, we compare our methods with several recent state-of-the-arts: 1) Basic methods: TA-DCL~\cite{zhang2023triplet} with intra-pool contrastive learning, TS-Former~\cite{zhu2022two}, MLSL-Net~\cite{yi2022multi}, C-Tran~\cite{Lanchantin_2021_CVPR}, and Q2L~\cite{liu2021query2label}), 2) Causality learning-based methods: CCD~\cite{liu2022contextual}, IDA~\cite{liu2022causality}, and Multilevel Causality (ML-C)~\cite{cui2024multilevel}. Mean average precision (mAP), class-wise recall (CR), class-wise F1 score (CF1), overall recall (OR), and overall F1 score (OF1) are used as evaluation metrics. 
\\
\textbf{Implementation Details.} We train our IBCA model on an NVIDIA Tesla A40 GPU. A hybrid ViT Backbone (ResNet-50+ViT-B$\_$16)~\cite{touvron2021training} with multiple class tokens is employed in our experiments, and all ViT-based methods share this backbone. Adam optimizer with an initial learning rate of $1e-4$ is adopted for a batch size of $64$. The training epoch is set to 200, and $\beta$ in $\mathcal{L}_{VIB}$ is set to 0.001. During inference, the mean average of patch-token and class-token-based predictions is used for the final prediction. Our code is available on GitHub \footnote{https://github.com/rabbittsui/IBCA}.
\begin{table*}[t]
\centering
\caption{Experiment results in comparisons with SOTAs on MuRed and Endo.}
\begin{tabular}{c|ccccc|ccccc} \hline
\multirow{2}{*}{Models} & \multicolumn{5}{c|}{MuRed} & \multicolumn{5}{c}{Endo} \\ \cline{2-11}
 & {CR$\uparrow$} & {CF1$\uparrow$}&  {OR$\uparrow$} & {OF1$\uparrow$} & {mAP$\uparrow$}& {CR$\uparrow$} & {CF1$\uparrow$}&  {OR$\uparrow$} & {OF1$\uparrow$} & {mAP$\uparrow$} \\
 \hline
CTRAN~\cite{Lanchantin_2021_CVPR}       & 39.44  & 40.26   & 51.58  & 60.12  & 59.11 & 53.36  & 54.99    & 51.42  & 53.43  & 61.28 \\
Q2L~\cite{liu2021query2label}               & 61.32  & 48.28   & 71.93  & 61.84  & 61.42 & 61.48  & 65.37   & 61.32  & 64.52  & 67.95\\
TS-Former~\cite{zhu2022two}          & 39.16  & 44.99    & 54.04  & 62.86  & 61.97  & 39.16   & 44.99   & 54.04  & 62.86  & 61.97\\
MLSL-Net~\cite{yi2022multi}    & 52.04  & 54.91   & 64.91  & 66.43  & 60.35 & 62.98  & 67.53   & 65.09  & 68.15  & 71.37 \\
TA-DCL$\ast$~\cite{zhang2023triplet}  & 46.05  & 50.50    & 52.63  & 60.48  & 56.54   & 63.92  & 64.86  & 60.38  & 62.29  & 64.34 \\
\hline
IDA~\cite{liu2022causality}   & 55.74  & 58.18    & 63.86  & 67.41  & 63.42  & 64.71  & 70.04    & 65.09  & 69.52  & 74.96\\
CCD~\cite{liu2022contextual}    & 60.39  & 57.15  & 64.21  & 66.79  & 64.90 & 61.59  & 70.40    & 62.26  & 69.47  & 74.03\\
ML-C~\cite{cui2024multilevel}  	&59.69	&59.59		&64.91	&69.16	&68.41 &	60.95&	66.48&	63.21&	67.17&	74.21\\
Ours    & \textbf{66.74}  & \textbf{59.90}   & \textbf{72.63}  & \textbf{69.22}  & \textbf{73.43} & \textbf{66.18}  & \textbf{72.05} & \textbf{66.04}  & \textbf{70.00}  & \textbf{76.38 }\\
\hline
\end{tabular}
\label{table1}
\end{table*}
\subsection{Comparison with State-of-the-art Methods}
We validated the effectiveness of the proposed IBCA on the MuRed and Endo datasets in Table~\ref{table1}. It is observed that the causality learning-based method outperforms the basic methods on most metrics, due to its efficient de-confounding of the contextual bias by different causal intervention strategies. However, our method incorporates the information bottleneck theory into the causal attention of ViT, and possesses superior capability on both datasets. Compared to ML-C~\cite{cui2024multilevel}, our method brings improvements with increments of 7.05\% in CR, 7.72\% in OR, and 4.95\% in mAP for MuRed, respectively. Compared to IDA~\cite{liu2022causality}, our method brings improvements with increments of 1.74\% in CR, 2.01\% in CF1, and 1.42\% in mAP for Endo, respectively. These results demonstrate the effectiveness of our approach in learning minimal redundancy and the most discriminative causal class-aware features for different labels.
\begin{table*}[t]
\centering
\caption{Ablation study of our proposed modules on MuRed and Endo.}
\begin{tabular}{c|ccccc|cccccc} 
\hline
\multirow{2}{*}{Methods} & \multicolumn{5}{c|}{MuRed} & \multicolumn{5}{c}{Endo} \\ 
\cline{2-11}
 & {CR$\uparrow$} & {CF1$\uparrow$}&  {OR$\uparrow$} & {OF1$\uparrow$} & {mAP$\uparrow$}& {CR$\uparrow$} & {CF1$\uparrow$}&  {OR$\uparrow$} & {OF1$\uparrow$} & {mAP$\uparrow$} 
  \\ 
  \hline
Basic &
59.26  & 57.80  &  63.16  & 68.77  & 72.81 & 63.00	& 68.44    & 64.62	 & 69.02  & 75.34
\\
Single VIB &
66.11 &	58.71	& 68.42	& 67.36  & 73.28 & 63.65 &	68.63	 &	65.57&	69.50	&76.14 \\ 
GMM VIB&
 65.11   &  59.64    &  69.82   &  \textbf{69.82 } 
 &  72.59  &  65.70   &  70.62  &  65.57   &  \textbf{70.03}   &  76.38  \\
 Ours&    
\textbf{66.74}  & \textbf{59.90}   & \textbf{72.63}  & 69.22  & \textbf{73.43} & \textbf{66.18}  & \textbf{72.05} & \textbf{66.04}  & 70.00  & \textbf{76.38 }\\
 \hline
\end{tabular}
\label{table2}
\end{table*}
\subsection{Ablation Study}
\begin{figure}[t]
\includegraphics[width=\textwidth, trim=10 0 0 0, clip]{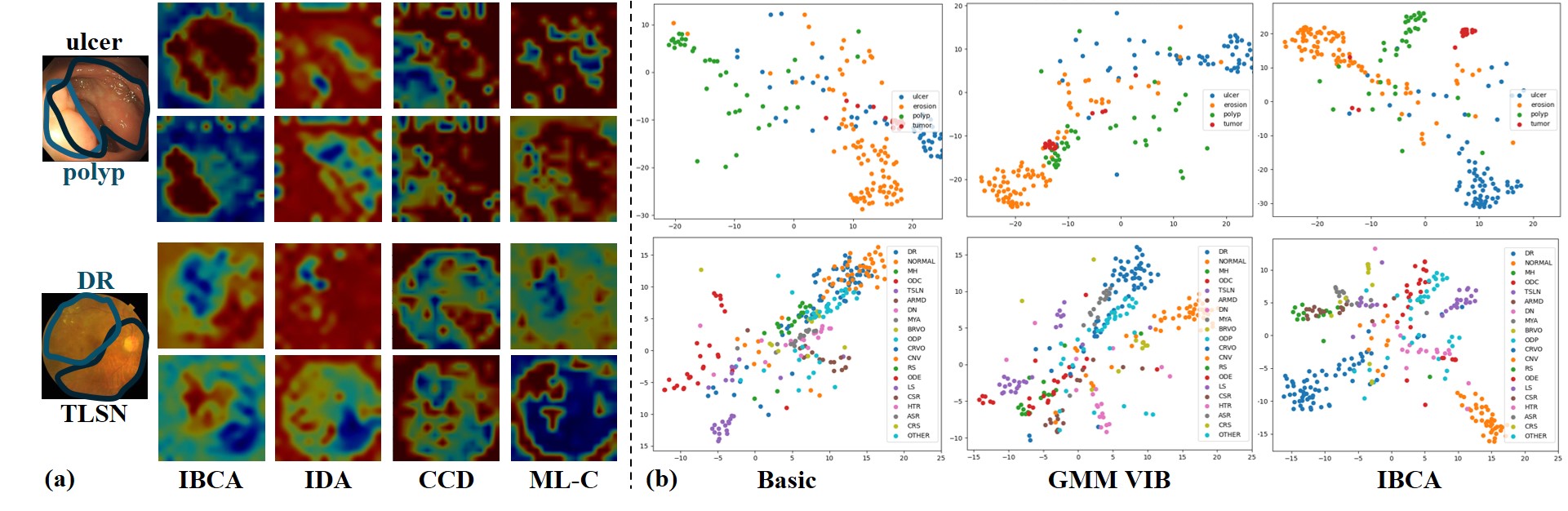}
\caption{Visualization results demonstrated the effectiveness of our method. (a) Visualization of class-specific spatial attention from different methods. (b) T-SNE results of class-specific features from different settings in ablation study.}
\label{fig3}
\end{figure}
An ablation study is conducted to evaluate the effectiveness of Gaussian mixture multi-label spatial attention, VIB loss, and contrastive enhancement attention in our method. Results are shown in Table~\ref{table2}. In the first column of Table~\ref{table2}, Basic refers to removing all three key components, with a fully connected layer to learn spatial mapping in the patch-token path. Single VIB substitutes the Gaussian Mixture Model (GMM) distribution with a single Gaussian distribution without $\mathcal{L}_{s}$, and GMM VIB represents our GMM-based spatial attention with VIB without $\mathcal{L}_{s}$. Compared to the basic architecture, both Single VIB and GMM VIB show significant improvements, particularly on MuReD, with increases of 6.85\% and 5.26\% in CR and OR, respectively. This suggests the effectiveness of the IB in filtering out task-irrelevant information, particularly when the number of predefined class labels is smaller. Notably, GMM VIB outperforms Single VIB across most metrics. For MuReD, CF1 improves by 0.93\%, OR by 1.40\%, and OF1 by 2.46\%; for Endo, CR increases by 2.05\%, CF1 by 1.99\%, and OF1 by 0.47\%. These results highlight the superiority of our GMM-based multi-label spatial attention in reducing class-irreverent features. Finally, $\mathcal{L}_{s}$ incorporates the IB constraint to reduce the noise factor from causal intervention, achieving the best performance in most metrics.

To intuitively demonstrate the effectiveness of our IBCA, we visualize the class-specific spatial attention with correct predictions of different causal-based methods. The corresponding lesion regions are highlighted in Fig.\ref{fig3} to investigate the interpretability of learned attention maps. It is observed that the task-irrelevant information and spurious correlation are contained in the attention maps from IDA and CCD, while ML-C tends to contain some task-irrelevant information with a causal part. In comparison to these methods, our IBCS captures more comprehensive lesion regions, especially the emphasis on the retinal vein lesions in fundus images, which contributes to better interpretability for diagnosis. In addition, t-SNE~\cite{van2008visualizing} is applied to visualize the distribution of patch-token-based class-specific features, as shown in Fig.~\ref{fig3}. Clear boundaries between classes are observed on both the Endo and MuRed datasets, demonstrating that our method effectively discriminates class-specific features by leveraging GMM-based spatial attention.
\section{Conclusion}
To the best of our knowledge, our IBCA is the first to incorporate IB into causality learning, enabling the learning of discriminative causal class-specific attention for multi-label medical image recognition. IBCA effectively learns IB-based Gaussian mixture multi-label spatial attention to mitigate the influence of task-irrelevant information. This Gaussian-mixed attention is further integrated into MHFS-based causal intervention to separate the redundancy noise from causal class-specific attention for class-specific feature learning. Extensive experiments on two public datasets demonstrate that our method outperforms existing methods by remarkable margins and shows better interpretation ability.

\begin{credits}
\subsubsection{\ackname} This study was funded by Shandong Natural Science Foundation under Grant ZR2024QF209.

\subsubsection{\discintname}
The authors have no competing interests to declare that are
relevant to the content of this article.
\end{credits}

\bibliographystyle{splncs04}
\bibliography{paper-4323}
\end{document}